%% file: main.tex
\documentclass[12pt]{l4dc2021}

\usepackage{times}
\usepackage{titlesec}
\usepackage{wrapfig}
\usepackage{xcolor}
\usepackage{xspace}

\title[Visually Guided Latent Actions]{Learning Visually Guided Latent Actions for Assistive Teleoperation}

\author{%
 \Name{Siddharth Karamcheti} \Email{skaramcheti@stanford.edu} \\
 \addr Department of Computer Science \\
 Stanford University
 \AND
 \Name{Albert J. Zhai} \Email{albertz@caltech.edu} \\
 \addr Department of Computing \& Mathematical Sciences \\
 California Institute of Technology
 \AND
 \Name{Dylan P. Losey} \Email{losey@vt.edu} \\
 \addr Department of Mechanical Engineering \\ 
 Virginia Polytechnic Institute and State University
 \AND
 \Name{Dorsa Sadigh} \Email{dorsa@stanford.edu} \\
 \addr Department of Computer Science \\
 Stanford University
}

\hypersetup{
    colorlinks=true,
    linkcolor=orange,
    filecolor=magenta,      
    urlcolor=orange,
    citecolor=orange,
}

\begin{document}
\maketitle

\begin{abstract}
\input{sections/00_abstract}
\end{abstract}


\section{Introduction}
\label{sec:introduction}
\input{sections/01_introduction}

\section{Related Work}
\label{sec:related-work}
\input{sections/02_related-work}

\section{Formalism for Visually Guided Assistive Teleoperation}
\label{sec:formalism}
\input{sections/03_formalism}

\section{Visually Guided Latent Actions}
\label{sec:vla}
\input{sections/04_vla}

\section{Robosuite Environments}
\label{sec:robosuite-env}
\input{sections/05_robosuite}

\section{Simulation Experiments}
\label{sec:sim-experiments}
\input{sections/06_simulation-experiments}

\section{User Study -- Teleoperating a 7-DoF Arm}
\label{sec:user-study}
\input{sections/07_user-study}

\section{Conclusion}
\label{sec:conclusion}
\input{sections/08_conclusion}

\acks{This work was graciously supported by the NSF (\#2006388) and the Future of Life Institute Grant RFP2-000. S.K. is additionally supported by the Open Philanthropy Project AI Fellowship.}
\bibliography{references}

\end{document}

%% file: sections/00_abstract.tex
It is challenging for humans -- particularly those living with physical disabilities -- to control high-dimensional, dexterous robots. Prior work explores learning embedding functions that map a human's low-dimensional inputs (e.g., via a joystick) to complex, high-dimensional robot actions for assistive teleoperation; however, a central problem is that there are many more high-dimensional actions than available low-dimensional inputs. To extract the correct action and maximally assist their human controller, robots must reason over their \textit{context}: for example, pressing a joystick down when interacting with a coffee cup indicates a different action than when interacting with knife. In this work, we develop assistive robots that condition their latent embeddings on visual inputs. We explore a spectrum of visual encoders and show that incorporating object detectors pretrained on small amounts of cheap, easy-to-collect structured data enables i) accurately and robustly recognizing the current context and ii) generalizing control embeddings to new objects and tasks. In user studies with a high-dimensional physical robot arm, participants leverage this approach to perform new tasks with unseen objects. Our results indicate that structured visual representations improve few-shot performance and are subjectively preferred by users.\footnote{Code for collecting demonstrations on a Franka Emika Panda Arm, pretraining visual encoders, and training visually guided latent actions models is available here: {\fontfamily{txtt}\selectfont \url{https://github.com/Stanford-ILIAD/vla}. }}

%% file: sections/01_introduction.tex
\begin{figure*}[t]
    \centering
    \includegraphics[width=\linewidth]{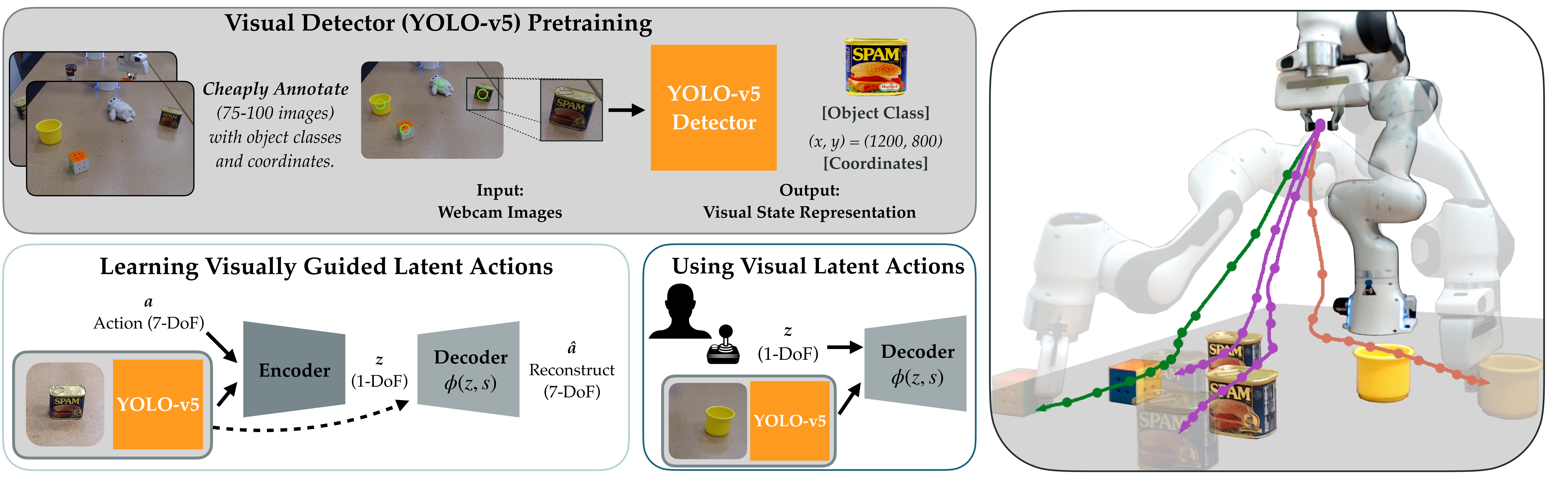}
    \vspace{-25pt}
    \caption{\small{Visually Guided Latent Actions. Starting from a small, annotated dataset of objects, we train an \textit{object detection} model -- YOLO-v5 \citep{redmon2016yolo} -- to encode visual state representations. Using these representations, we train a conditional autoencoder (CAE) from a small set of demonstrations, encoding high DoF actions to a 1D latent space. During teleoperation, this visually guided decoder provides an intuitive 1-DoF control interface for human users.}}
    \label{fig:front}
    \vspace*{-7mm}
\end{figure*}

Nearly one million American adults living with physical disabilities require help when performing everyday tasks \citep{taylor2018americans}. Assistive robots -- such as wheelchair mounted robot arms -- are a promising solution, but are difficult to control. Consider operating a $7$ degree of freedom (DoF) arm to reach out and move a cup (Figure \ref{fig:front}). We first must guide the robot to reach our high-level goal, then carefully orchestrate all its joints to manipulate the cup. Because of their physical limitations, our target users cannot show the robot how to do this task directly: instead, they rely on \textit{low-dimensional} control interfaces (e.g., joysticks, sip-and-puff devices) \citep{argall2018autonomy}.

Prior work tackles this by learning an \textit{embedding function} that maps low-dimensional human inputs to high-dimensional robot actions \citep{losey2020latent, jeon2020sharedlatent}. Here, an able-bodied user kinesthetically guides the robot through an assortment of tasks. The robot embeds these high-dimensional demonstrations to a low-dimensional latent action space, then decodes from this latent space to reconstruct the demonstrated actions. Any user can use the resulting system to control the robot using the low-dimensional latent space (i.e., treating joystick inputs as \textit{latent actions}). 

For this approach to work, the robot must reason over its \textit{context}. Consider our example in Figure \ref{fig:front}: when the human reaches for the cup, the robot should maneuver its gripper under the lip of the cup. But what if the human instead reaches for the SPAM? Now, that same motion does not make sense; instead, the robot should attempt to scoop out some food. To map users' low-dimensional inputs to meaningful actions, a robot \textit{must} consider its current state, and the \textit{types} and \textit{locations} of nearby objects. While critical, prior work avoids the challenge of considering various contexts through \textit{hard-coding} object locations; a solution that clearly fails to scale to the uncertainty and dynamics of the real world -- for example, imagine having to remap and specify the location of items in your fridge every time you buy groceries. Moving forward, we need robots that leverage visual inputs, applying methods that are capable of accurately \textit{detecting} and \textit{localizing} objects.

We introduce a framework for visually guided latent action controllers for assistive teleoperation. Under our approach, robots leverage the output of pretrained object detectors to form robust representations of context. We show that our robot with access to these structured representations is able to learn and generalize from limited human demonstrations to a meaningful latent action space -- one suited for intuitive, low-dimensional control. Overall, we make the following contributions:

\smallskip
\noindent \textbf{Formalizing Visual Latent Actions.} We introduce a framework for incorporating visual state representations into latent action learning for assistive teleoperation.

\smallskip
\noindent \textbf{Evaluating a Spectrum of Visual Input Encoding Strategies.} We consider a variety of visual models, from end-to-end models that directly process pixel inputs, to structured representations that rely on modular object detection architectures. We investigate the sample efficiency and few-shot transfer abilities of these models over a broad suite of experiments in simulation.

\smallskip
\noindent \textbf{Conducting a User Study.} We test these approaches with a user study, where real participants teleoperate an assistive robot arm that must rely on visual inputs. Our results show that our detector-backed latent action controller significantly outperforms other control strategies in terms of success rate, as well as subjectively, in terms of naturalness, intuitiveness, and willingness to reuse in future.

%% file: sections/02_related-work.tex
We build our approach atop a long-line of work around dimensionality reduction for assistive teleoperation \citep{ciocarlie2009hand, jain2019probabilistic} -- including prior work on visually-guided manipulation and learned latent actions \citep{losey2020latent, jeon2020sharedlatent, li2020intuitive}.

\paragraph{Dimensionality Reduction for Assistive Teleoperation.}
In assistive teleoperation, the dimensionality of the human control interface (for example, a wheelchair-attached joystick) does not usually match up with the degrees of freedom (DoFs) of the robot (in many cases 6+), increasing control difficulty. Teleoperating these robots with devices like joysticks or sip-and-puff devices requires frequent ``mode-switching,'' using multiple sets of joystick axes to control the linear or angular mode of the end-effector. This is time-consuming, challenging, unintuitive, and fails to provide the user with fine-grained control \citep{wilson1996relative}. Conversely, other teleoperation methods that measure and replicate users' limb movement \citep{conti2014interface, laghi2017teleimpedance, laghi2020unifying} often require access to specialized hardware and are difficult to use by users with disabilities. 

Learned latent actions \citep{losey2020latent} tackles this issue by using conditional autoencoders (CAEs) \citep{doersch2016tutorial} to map from low-dimensional inputs to high-dimensional actions, providing users with intuitive, task-conditional control. Prior work combined latent actions with shared autonomy for increased precision \citep{jeon2020sharedlatent}, as well as personalized the mapping of the learned latent space to individual users \citep{li2020intuitive}. However, \textit{these methods assume a fixed environment with ground-truth access to the environment state} -- unrealistic assumptions if we want to deploy robots in new, dynamic scenarios. 

\paragraph{Visually-Guided Manipulation and Control.}
Recent work leverages advances in deep learning and computer vision to facilitate precise and robust manipulation. These approaches often perform pose estimation of various objects of interest (after supervised training) and plan the respective actions \citep{wang2019learning, feng2019robot, park2020toward}. Additional work has shown progress in learning visual representations end-to-end using task reward signals \citep{lai2012detection, levine2016end2end, zhu2017target, yu2019unsupervised}, or through self-supervised encoding of play behavior \citep{wang2019learning, lynch2019learning}. Building off these advances, our work uses pose-based visual representations of the environment to condition low-dimensional embedding spaces for high-DoF robot actions, allowing for generalization to unseen object layouts.

%% file: sections/03_formalism.tex
\paragraph{Motivating Example.} 
Figure \ref{fig:front} presents a workspace with several objects positioned in front of a Franka Emika Panda robot arm. Each user teleoperates the robot to perform a series of \textit{tasks}: in this case pushing objects around the table. In real world settings, a user would like robots to perform goal-directed behavior (e.g., turning doorknobs or retrieving items from the fridge); similarly, in this work, we assume each object is associated with its own unique task. For example (Figure \ref{fig:front}), the SPAM is associated with pushing south, while the Rubik's Cube is associated with pushing south-east. Success is defined by whether the user is able to teleoperate the robot to guide the object a set distance in the desired direction.

We need an assistive teleoperation method that can handle disparate tasks across multiple object categories and varying positions. One path towards this goal is through leveraging visual inputs; given an image of a workspace, we 1) identify the types of objects on the workspace (\textit{detection}), and 2) identify the corresponding positions (\textit{localization}). Here, we describe the formalism we use for learning an effective and intuitive visually guided assistive teleoperation interface.

\paragraph{Formalism.}
We assume the visual input for our assistive interface is provided by an RGB camera mounted at a fixed point relative to the robot (as in Figure \ref{fig:front}). A robot manipulation task is formulated as a fully-observable, discrete time Markov Decision Process (MDP) $\mathcal{M} = (\mathcal{S}, \mathcal{A}, \mathcal{T}, \mathcal{C}, R, \gamma, \rho_0)$. The state space $\mathcal{S}$ is represented by the cross-product of robot joint states and visual observations $\mathbb{R}^m \times \mathbb{R}^{H \times W \times 3}$, where $m$ denotes the robot-DoF, and ($H$, $W$) denote the resolution of the image observation. $\mathcal{A} \subseteq \mathbb{R}^m$ is the robot action space, $\mathcal{T}(s, a)$ is the transition function, $\gamma \in [0, 1]$ is the discount factor, and $\rho_0$ is the initial state distribution.

Let $\mathcal{C}$ be the discrete set of object categories that may appear (for example, the SPAM and Cup from Figure \ref{fig:front}). The environment contains a set of objects $\mathcal{O} \subseteq \mathcal{C}$; at each timestep $t$, the human teleoperator has a goal object $o_t \in \mathcal{O}$ that they wish to manipulate (for example, the SPAM in Figure \ref{fig:front}); each object has a unique action associated with it that dictates reward (we want to push the SPAM \textit{south}), where reward $R(s_t, o_t)$ is a function of the object $o_t$ and current state $s_t$. Building a successful interface for assistive teleoperation involves learning components capable of both \textit{detecting} and \textit{localizing} objects, in addition to producing the context-specific actions necessary to perform the requisite motion. Our core contribution is processing visual inputs to guide the learning of low-dimensional representations for generalizable and intuitive assistive teleoperation.

%% file: sections/04_vla.tex
We leverage \textit{latent actions} to learn visually guided low-dimensional control interfaces for assistive teleoperation. Latent actions is an umbrella term, referring to a class of approaches that learn low-dimensional representations of high-dimensional actions through dimensionality reduction \citep{jonschkowski2015statereps, lynch2019learning, losey2020latent}. In this work, we take the approach proposed by \citet{losey2020latent} that uses conditional autoencoders (CAEs) \citep{doersch2016tutorial} for learning latent actions. Given a set of demonstrations consisting of state-action pairs $(s, a)$, the robot jointly encodes the state $s$ and high-DoF action $a$ into a low-dimensional embedding space, producing a latent embedding (i.e., latent action) $z$. Conditioned on $z$ as well as the original state representation $s$, the robot then attempts to learn a decoder $\phi(z, s) : \mathcal{Z} \times \mathcal{S} \mapsto \mathcal{A}$ that \textit{reconstructs} the original high-DoF action $a$. 

This approach works well in practice -- conditioning the output of $\phi$ on the state representation $s$ allows the robot to learn a low-dimensional embedding space that captures the mode of the demonstrations \textit{conditioned on the context}, learning smooth, intuitive mappings. The secret, however, is in the structure of the state representation $s$: any representation that is too unstructured prohibits the decoder $\phi$ from fitting the demonstration distribution, as there are too many contexts to identify meaningful patterns and learn representative low-dimensional actions. Prior work circumvents this problem by assuming fixed environments or access to the ground truth environment state -- which is unrealistic when deploying assistive robots in dynamic, real-world settings. Addressing this problem of learning a good state representation is particularly challenging given visual inputs -- what is the best way to encode observations? In the next section, we explore three strategies with different structural biases for \textit{detecting} object types and \textit{localizing} object positions, with the hope of learning a visual state representation that can be used as a conditioning input for learning latent actions.

\subsection{Reliable and Efficient Detection and Localization}
\label{sec:detect-localize}

We study three strategies for mapping visual inputs (raw RGB images) to state representations.

\paragraph{Fully End-to-End Approaches.}
We first look at training a convolutional neural network (CNN) encoder from scratch solely using visual observations paired with demonstrations in our dataset. We pass the RGB image through a CNN architecture with max-pooling, recovering a dense feature vector $f$. The fused robot \& environment state representation $s$ is a 7-DoF joint state concatenated with this dense feature vector: $s = s_{\text{joint}} \oplus f$. We learn this CNN encoder jointly with the latent actions CAE.

\paragraph{Learning Localization.}
Jointly learning \textit{detection} as well as \textit{localization} in the scope of a single CNN encoder may be difficult given limited examples (in our case, no more than 10 demonstrations for a given object class). To test this, we experiment with a ``localization-only'' model provided with the ground-truth object class by an oracle prior to teleoperation. We use an identical CNN encoder as the end-to-end approach to learn how to \textit{localize} objects, and concatenate the final dense ``localization'' vector $g$ with a one-hot encoding of the object type $c$. The fused robot \& environment state representation $s$ is: $s = s_{\text{joint}} \oplus g \oplus c$.

\paragraph{Structured Representations \& Leveraging Static Data.}
Both prior approaches are \textit{uninterpretable}; it is not clear what the CNN encoders are learning or what information they preserve. This is particularly concerning when teleoperating a robot arm in safety critical settings. To combat this, we propose using a third strategy with a strong structural inductive bias: predicting object classes and $(x, y)$ coordinates directly from an RGB image. We use the YOLO-v5 architecture \citep{redmon2016yolo, jocher2020yolov5}, due to its detection accuracy and speed on existing benchmarks. We found that we needed only 75-100 labeled images of 2-4 objects each to train a reliable detector.

The structural inductive bias present in YOLO-v5 directly addresses both \textit{detection} and \textit{localization}. Crucially, after pretraining YOLO-v5 on our static dataset, \textit{we hold it fixed during CAE training}, theoretically allowing for more stable and reliable training dynamics than with the prior approaches. The fused state representation is given as the joint state concatenated with the detected class one-hot vector $c$, and the predicted $(x, y)$ coordinates from YOLO-v5: $s = s_{\text{joint}} \oplus c \oplus [x, y]$.

\subsection{Few-Shot Adaptation}

Reliably detecting and localizing multiple objects enables robots to learn a wide-variety of object-specific behaviors. Furthermore, as the latent actions model is shared across all objects (and thus, all tasks), there is an opportunity for positive transfer and robust few-shot learning when faced with brand new tasks or objects; as long as there is \textit{plausible} shared structure between the new task and a task seen at training (e.g., pushing south at train to push south-west at test), we might expect the latent actions model to adapt with fewer demonstrations than if it were learning from scratch. We perform experiments to this effect, evaluating how latent action models degrade as we 1) decrease the number of provided demonstrations for an unseen task (in a ``few-shot setting''), and 2) increase the distance between the unseen and seen behaviors.

%% file: sections/05_robosuite.tex
We first perform experiments with a simulated Franka Emika Panda robot arm (7-DoF) to evaluate the various visual encoding schemes and few-shot potential described in Section \ref{sec:vla}. We use the Surreal Robotics Suite (Robosuite) \citep{fan2018surreal}, a simulation toolkit based off the MuJoCo physics engine \citep{todorov2012mujoco} to render environments. 

\paragraph{Setup.} The robot state $s_{\text{joint}} \in \mathbb{R}^7$ consists of joint positions, while the action $a \in \mathbb{R}^7$ corresponds to joint velocities. The robot's next state is given by $s_{\text{joint}}' = s_{\text{joint}} + a \cdot dt$, where $dt$ corresponds to step size (held fixed at 1.0). Our visually guided latent action models are trained to map to latent actions $z \in \mathbb{R}^1$ (1-DoF), and we use the corresponding decoder $\phi$ as our visually guided assistive controller. In simulation, we use a planner to generate oracle demonstrations for individual tasks. When accomplishing tasks, we extract a single image at the beginning of the interaction episode to feed to our visual encoders to obtain the environment state information \textit{for the entire episode}; constantly running the image processing loop significantly impacts latency, with minimal benefit.

\paragraph{Simulated Teleoperator.}
In lieu of a real human using a 1-DoF joystick to control the robot with our visually guided latent actions model, we build a \textit{simulated teleoperator} that greedily searches over a discretized range in $\mathbb{R}^1$ for the action that minimizes projected distance to a target \textit{waypoint} in joint space $w_{\text{joint}} \in \mathbb{R}^7$. In other words, given our visually guided latent action decoder $\phi$, our robot's joint state $s_{\text{joint}}$, our fused robot \& environment state $s$ (computed separately for each encoding strategy, as described in Section \ref{sec:detect-localize}), and action step size $dt$ we compute the locally-optimal latent action $z^*$ to take as: $z^* = \text{argmin}_{z \in \mathcal{Z}} ||w_{\text{joint}} - (s_{\text{joint}} + \phi(z, s) dt)||_2$.

Most of the tasks in our evaluation are pushing tasks; as such, they cannot be captured with a single waypoint $w_{\text{joint}}$ (with just the final point, the robot would just try to navigate directly there, without interacting with the object). To mitigate this, we sample 2 waypoints, $w_{\text{joint}}^{\text{pre}}$ and $w_{\text{joint}}^{\text{post}}$ denoting the \textit{pre-push} (prior to making contact with the object) and \textit{post-push} (after completing the push) points respectively. We use timeouts for switching between waypoints; for the first $k$ seconds we optimize our choice of latent actions $z^*$ with respect to the first waypoint (we use $k = 15$ seconds), then optimize $z^*$ with respect to the second for all future timesteps. We emphasize defining such intermediate waypoints is only to aid our \textit{simulated} teleoperator that can later be used for evaluation of our method \textit{in Robosuite}. In general (and as evidenced by our User Study in Section \ref{sec:user-study}), human users use our visually guided assistive controller without a task breakdown.

\paragraph{Pretraining YOLO-v5 on Simulated Workspaces.}
To build our initial dataset for pretraining our YOLO-v5 detector in simulation, we sample 1000 random configurations of individual objects (including the few-shot target type) on the workspace, with the robot joint state also sampled randomly, and programmatically annotate them with $(x, y)$ coordinate and object class.

Note that to generalize to unseen objects, we require retraining our detector given annotated images of the novel object. As part of our experiments in Section \ref{sec:fewshot}, we evaluate our visually guided model's ability to generalize to new tasks given a previously unseen object; for those experiments we \textit{include} static images of the ``unseen'' object in our YOLO-v5 dataset (part of the 1000 random configurations above); however, \textit{we do not train on any demonstrations for the given object ahead of time} (making these objects effectively ungrounded). This is an approximation of how one might deploy a visually guided system in practice (perhaps after pretraining on several objects at scale). 

\subsection{Evaluating Visual Encoding Strategies}
\label{sec:base}

To evaluate the relative performance of our visual encoding strategies (End-to-End, Localization-Only, and YOLO-v5), we select 4 different objects (\textit{bread, soda can, milk carton, and glass bottle}) and collect 10 \textit{base} demonstrations for each of them, with each object assigned a unique pushing task (\textit{pushing east, south, south-west, and west}). We evaluate our strategy by looking at performance relative to the number of training demonstrations provided, looking at model performance after training on 1 demonstration per object, then 2 demonstrations, and so on, up through the full dataset of 10 demonstrations per object (10 iterations total). After training each visually guided latent actions model with a given encoder on a dataset of demonstrations, we evaluate performance by using the Simulated Teleoperator on a set of 10 validation tasks for each of the 4 objects. 

\subsection{Few-Shot Generalization: Efficiently Learning Latent Actions for Unseen Objects}
\label{sec:fewshot}

To evaluate the few-shot/transfer learning potential of our approach, we: 1) Train a no-transfer variant of the latent actions model, where the demonstrations only consist of the unseen class (the few-shot demonstrations), and 2) Train a transfer variant of the latent actions model where the demonstration dataset consists of all 40 demonstrations from each the base classes \textit{in addition to} demonstrations from the unseen class. For all simulation experiments, the unseen object is a \textit{cereal box}. Critically, we look at varying the ``difficulty'' (behavior novelty) of the unseen tasks as follows:

\paragraph{\textsc{Seen}.} Given a new object, try and replicate a task from the base set. For example, one of the the base tasks is \textit{Push the bread south}, so our \textsc{Seen} task is \textit{Push the cereal box south}.

\paragraph{\textsc{Near}.} Given a new object, perform a task slightly different from the base set. Base tasks include pushing \textit{south, south-west, west, and east}, so our \textsc{Near} task is \textit{Push the cereal box south-east}.

\paragraph{\textsc{Far}.} Given a new object, perform a task different from any of the base tasks. Our \textsc{Far} task is \textit{Rotate around the cereal box in a circle}, a motion that is not represented in the base set.

%% file: sections/06_simulation-experiments.tex
\paragraph{Model Details.}
In all of our experiments, both the encoder/decoder in our latent actions CAE are Multi-Layer Perceptrons (MLPs) with 2 hidden layers. The CAE has a single-node bottleneck (the learned latent action space is 1-dimensional), and is trained to minimize the $L_2$ reconstruction loss on the dataset of robot actions. State-action pairs for training are computed from pairwise joint state differences for subsequent states (within a fixed window) in a demonstration. To prevent overfitting, we augment states in our dataset by adding noise $\epsilon \sim \mathcal{N}(0, \sigma^2)$, where $\sigma = 0.01$.

\paragraph{Configuration \& Metrics.}
We optimize for \textit{Final State Error}, measured as the L2 Distance between the final joint state of the robot and the final waypoint provided to the teleoperator, after a time limit of 30 seconds. We average performance over 10 runs, sampling different subsets of demonstrations for each new run. We additionally perform qualitative analyses of the end-effector trajectories for each strategy to identify common failure modes and other patterns.

\paragraph{Hypotheses.}
We have the following three hypotheses:

\paragraph{H1.} \textit{Provided sufficient demonstrations, each of the three visual encoding strategies from Section \ref{sec:detect-localize} will learn accurate latent actions that accomplish the appropriate tasks for each object class.}

\paragraph{H2.} \textit{Of the three visual encoding strategies we evaluate, the YOLO-based detection model that produces the most structured state representation -- detected object class labels and coordinates -- will be the most sample efficient of the visually guided latent action models.}

\paragraph{H3.} \textit{Transfer learning will almost always be better than learning latent actions from scratch given a small dataset of unseen demonstrations.}

\subsection{Results \& Analysis}

\begin{figure*}[t]
    \centering
    \includegraphics[width=\linewidth ]{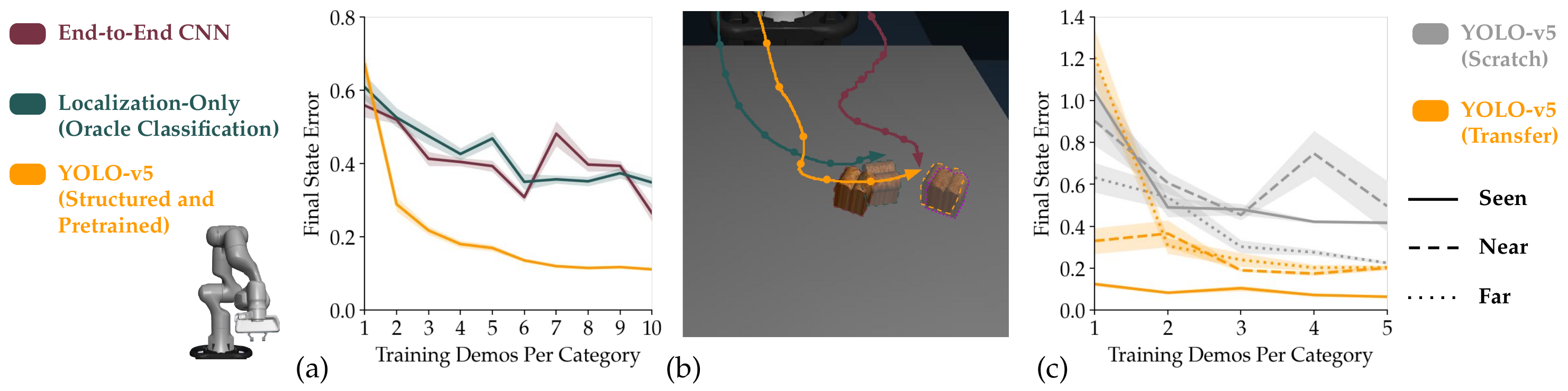}
    \vspace{-27pt}
    \caption{\small{Robosuite Simulation Experiments, with plots and visualizations investigating the sample efficiency of the various visual encoding strategies, and the benefit of transfer learning when trying to generalize to unseen objects and new tasks. Shaded regions denote standard error across runs.}}
    \label{fig:robosuite}
    \vspace*{-8mm}
\end{figure*}

\paragraph{Evaluating Visual Encoding Strategies.}

Figure \ref{fig:robosuite}a presents a graph of final state error vs. \# of demonstrations per object class for the three visual encoding strategies. Surprisingly, access to the ground-truth object class does not seem to provide the localization-only model with an advantage over the end-to-end model. Generally, while all visual encoding strategies improve as they are exposed to more demonstrations, the structured, pretrained representation output by the YOLO-v5 model leads to superior performance; with just 2 demonstrations, the YOLO-v5 model is able to achieve better final state error than either CNN-based model trained with 10. Figure \ref{fig:robosuite}b presents a qualitative take; even after 10 demonstrations, the end-to-end model fails to make contact with the object (the bread), whereas the localization-only model is able to successfully start pushing, but not far enough to make it to the target location. Only the YOLO-v5 latent actions model is able to smoothly push the object to the desired location.

\paragraph{Few Shot Generalization to Unseen Objects.}
Figure \ref{fig:robosuite}c depicts the transfer (orange) and non-transfer (gray) performance across the \textsc{Seen}, \textsc{Near}, and \textsc{Far} settings. Unsurprisingly, in the \textsc{Seen} setting, the transfer models are able to get incredibly low final state error starting with just 1 demonstration. With the \textsc{Near} results, transfer models start off with an advantage at just a 1 demonstration, and continue to outperform the non-transfer models all the way up through training on the full set of 5 demonstrations available. However, looking at the \textsc{Far} results, we see that with just a single demonstration, the non-transfer model is able to outperform the transfer model by a large margin. That being said, this margin goes away completely after adding a second demonstration, at which point the transfer models take the lead for the remainder of the demonstration budget.

%% file: sections/07_user-study.tex
\begin{figure*}[t]
    \centering
    \includegraphics[width=\linewidth]{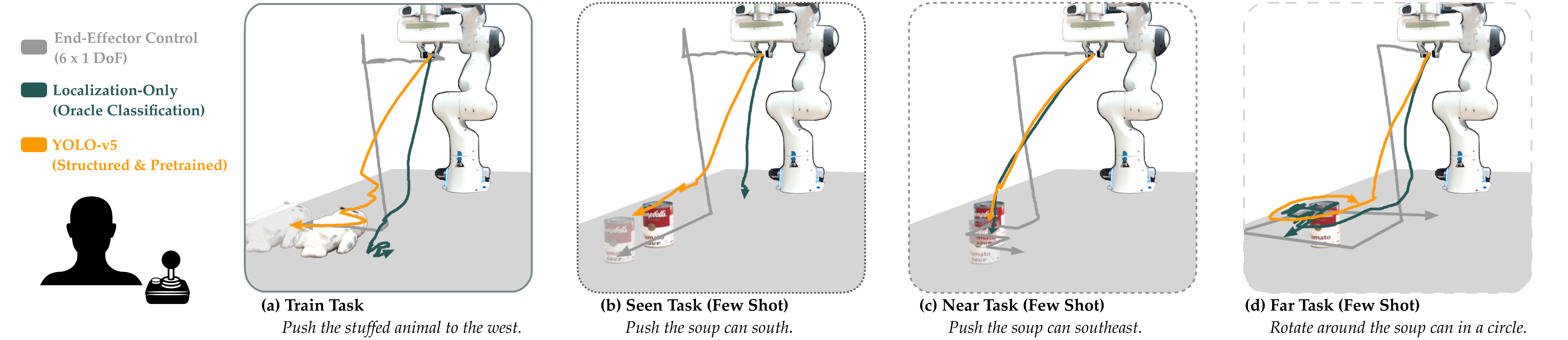}
    \vspace{-25pt}
    \caption{\small{Visualized trajectories from our user study. The Localization-Only model fails to reach the target object in most cases. Whereas the YOLO-v5 strategy allows for smooth and precise movement, the End-Effector trajectories are jagged and blocky, contraindicating its use for fine-grained tasks.}}
    \label{fig:user-trajectories}
    \vspace{-7mm}
\end{figure*}

In addition to the simulation experiments, we perform a user study that evaluates three different control strategies for accomplishing tasks with a 7-DoF robotic arm (also a Franka Emika Panda arm) with a 1-DoF joystick. Alongside our assistive approaches, we also include an end-effector control baseline, where users directly teleoperate the robot's motion by switching control modes \citep{herlant2016assistive}. 

\paragraph{Setup.}
To build a real-world visual latent actions model, we 1) collected an annotated image dataset of objects and their corresponding coordinates to train our YOLO-v5 Detector, and 2) collected demonstrations for the different tasks in our base setup. This included pushing in 4 directions -- south, south-west, west, and east. We additionally needed a smaller number of demonstrations for our few-shot generalization experiments; the same \textsc{Seen} (pushing south), \textsc{Near} (pushing south-east), and \textsc{Far} (circling an object) tasks used in simulation.

We needed only 100 images (90 train, 10 validation), with each image containing on the order of 2-4 objects to train an accurate YOLO-v5 detector in our lab workspace. Similar to the simulation experiments, we also collect static annotated detection data for our ``unseen object'' (a Campbell's Soup Can), to be used for our few-shot generalization experiments.

We collect demonstrations for our tasks kinesthetically, by manually backdriving the robot end-effector to make the appropriate pushing motions while recording the robot joint states. For our 4 base objects -- SPAM (south), Cup (east), Stuffed Animal (west), Rubik's Cube (south-west) we collect 9 demonstrations each; for each of our unseen-settings, we only collect 3 demonstrations.

We keep the CNN encoder and CAE model architectures identical to the simulation experiments, and use the same training hyperparameters and augmentation scheme, except for batch size, which was much larger than in simulation. Similar to prior setup, we only run the first webcam frame of each episode through our visual encoders; again, we saw no benefit in running the detection loop on each action.

\paragraph{Experiment Procedure.}
For the study, we recruited 9 volunteers that provided informed consent (4 female, ages $24.22 \pm 1.6$), with 8/9 participants having prior robot teleoperation experience.\footnote{Due to the COVID-19 pandemic, we could only recruit users with preapproved access to the robotics lab.} Participants used a 1-DoF joystick to control a 7-DoF robot arm with the 3 strategies above. Each participant completed 4 tasks $\times$ 2 trials for each evaluation mode (base object manipulation, and the three few-shot experiments). On exposure to a new model or object, participants were given an opportunity to practice; we reserved a single fixed point on the workspace that the participant could reset the object to as many times as they wanted. Then, after verbally agreeing to move on, we \textit{moved the object to a separate starting point on the workspace}, and had the users complete two trials back to back. We ran a qualitative survey that the participants filled out after each control strategy, where we asked questions about usability and intuitiveness, amongst others (see Figure~\ref{fig:user-stidu}). 

\subsection{Results \& Analysis.}
\begin{figure*}[t]
    \includegraphics[width=\textwidth]{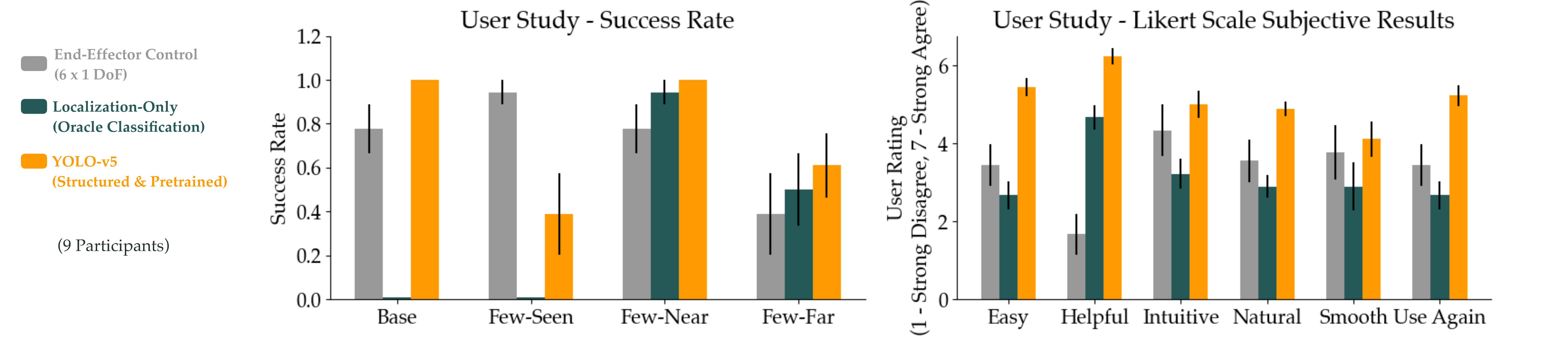}
    \vspace{-25pt}
    \caption{\small{User Study. \textbf{[Left]} Participants found YOLO-v5's structured representations crucial for better control, typically leading to high success rates on new and unseen tasks. \textbf{[Right]} Users preferred this structured approach across the board. Error bars denote std. error across the participant pool.}}    
    \label{fig:user-stidu}
    \vspace{-25pt}
\end{figure*}
Results are consistent with simulation; the YOLO-v5 latent actions outperform Localization-Only on all objective \textit{and} subjective metrics (Figure \ref{fig:user-stidu}), showing consistently higher success rates across the base (train) tasks, as well as the three few-shot tasks. Figure \ref{fig:user-trajectories} shows that the Localization-Only model's failures stemmed from a failure to \textit{localize} the object correctly, often going to a completely wrong point on the table (contributing to its 0\% success rate for certain tasks). However, it does succeed reliably on two of the harder tasks, indicating that learning to localize with an end-to-end model may still be viable, even with the complexity of the real-world.

More surprising is how well the End-Effector control strategy performed relative to the YOLO-v5 latent actions model, particularly in the \textsc{Seen} task. There are two explanations: the first is that given real-world demonstration data, 3 demonstration episodes are not enough for even the YOLO-v5 latent actions CAE to fit properly to the task. The second is that \textsc{Seen} is straightforward for the End-Effector control strategy, and requires only a few control inputs. Overall though, the YOLO-v5 model beats the End-Effector strategy on all subjective, Likert-scale metrics (Figure \ref{fig:user-stidu}), most notably in terms of ``ease of use,'' ``helpfulness,'' ``intuitiveness,'' and most importantly, ``whether the participant would use the given strategy again.'' Looking at the visualizations in Figure \ref{fig:user-trajectories} completes the story; most of the End-Effector controller's successes stem from very jagged, blocky motions, in contrast to the YOLO-v5 latent action model's smooth, natural-looking motion.

%% file: sections/08_conclusion.tex
\paragraph{Summary.} We developed a framework for incorporating visual state representations into latent action-based assistive teleoperation. Our simulations and results demonstrate that structured representations enable users to rapidly generalize learned control embeddings to new objects and tasks.

\paragraph{Limitations \& Future Work.} We built our system under the assumption that once detected, objects would stay mostly static; while this was an assumption that allowed us to make meaningful progress on
building a working version of this system, it is not a hard or fast constraint. Future work will look into expanding the existing models to deal with \textit{dynamic} objects, as well as more robust evaluations with users who already employ assistive robots in their everyday lives.